\documentclass[runningheads]{llncs}

\usepackage{accv}

\usepackage{accvabbrv}
\usepackage{graphicx}
\usepackage{booktabs}
\usepackage[table]{xcolor}
\usepackage{multirow}
\usepackage{wrapfig}

\usepackage{hyperref}
\usepackage{orcidlink}

\begin{document}

\title{ILLRAW: Illumination-Aware RAW Object Detection}

\titlerunning{ILLRAW: Illumination-Aware RAW Object Detection}
\newcommand{\equalcontrib}{\textsuperscript{*}}
\newcommand{\correspondingauthor}{\textsuperscript{\textdagger}}

\author{%
Xinzhe Wang\orcidlink{0009-0008-3044-1161}\inst{1}\equalcontrib
\and
Xiao Liang\orcidlink{0009-0008-1921-7732}\inst{1}\equalcontrib
\and
Xiaoli Zhang\orcidlink{0009-0007-9755-7217}\inst{3}
\and
Kang Ma\orcidlink{0000-0001-9996-8115}\inst{1,2}\correspondingauthor}

\authorrunning{X. Wang et al.}

\institute{Beijing Institute of Technology\and
State Key Laboratory of CNS/ATM, Beijing Institute of Technology\and
Independent Researcher\\[1ex]
\email{\textsuperscript{1}xinzwang123@gmail.com,
\textsuperscript{1}\{liangxiao,makang\}@bit.edu.cn\quad
\textsuperscript{3}z.xiaolilili@gmail.com}}

\makeatletter
\g@addto@macro\@thanks{%
  \footnotetext[1]{Equal contribution.}%
  \footnotetext[4]{Corresponding author.}}
\makeatother

\maketitle

\begin{abstract}
	

    Existing RAW-based detectors typically introduce learnable ISP modules to convert RAW inputs into sRGB-like formats to fit off-the-shelf detectors. 
    However, this RAW-to-sRGB paradigm decouples representation learning from the downstream detection task. Crucially, it forces the network to inherit human-centric visual priors while discarding intrinsic RAW advantages, such as linear sensor responses and high-bit-depth, minimally processed scene cues.
    In this paper, we propose \textbf{ILLRAW}, an \textbf{ILL}umination-aware end-to-end \textbf{RAW} object detection framework. 
    Instead of reconstructing visually plausible sRGB images, ILLRAW directly learns detection-oriented RAW features through a lightweight feature adapter inserted at the input stage of existing detectors.
    ILLRAW estimates an illumination mask and introduces channel-sensitive embeddings to model Bayer-channel responses. These features are fused by an illumination attention fusion module. Experiments show clear improvements on the diverse AODRaw benchmark and competitive performance on LOD and PascalRAW, with particularly strong results under dark exposure.

  \keywords{RAW object detection \and Illumination-aware Feature Adaptation \and end-to-end RAW detection}
\end{abstract}

\section{Introduction}
\label{sec:intro}

Object detection is a crucial and fundamental task in computer vision and has achieved significant progress~\cite{lin2014microsoft, shao2019objects365, ren2015faster, redmon2018yolov3, zhu2020deformable}. 
However, most existing object detection methods are developed on sRGB images, whose robustness under challenging imaging conditions is often overlooked~\cite{xu2023toward, li2024towards}. 
RAW images are the direct outputs of camera sensors~\cite{chen2018learning}, and their pixel values are linearly related to incident light intensity and preserve high-bit-depth radiometric information together with unprocessed sensor noise~\cite{zou2023rawhdr, cui2024raw}. 
Such sensor-level light measurements provide richer illumination cues than sRGB images, especially in dark regions, high-dynamic-range scenes, and areas close to saturation. 
By directly preserving sensor measurements and richer radiometric cues, RAW images provide a promising input space for robust object detection~\cite{wang2024multi, hong2021crafting, xu2023toward, li2024towards, cui2024raw} under challenging real-world conditions, such as high dynamic range scenes~\cite{xu2023toward, cui2024raw}, low-light environments~\cite{wang2024multi, hong2021crafting}, and adverse weather~\cite{li2024towards}.

In modern camera systems, RAW images are processed by the ISP and converted into sRGB images for human viewing. 
A typical ISP pipeline includes denoising, demosaicking, white balancing, gamma correction, and compression~\cite{ehret2019joint,zou2023rawhdr}. Although these operations improve visual quality, they inevitably reshape the original light distribution and may discard task-relevant details. 
For example, denoising suppresses sensor noise but can also remove fine structures~\cite{brooks2019unprocessing}, particularly in dark areas and near overexposed regions. 

To improve detection on RAW images, previous methods often couple object detectors with imaging systems by tuning the hyperparameters of the ISP or the camera pipeline~\cite{mosleh2020hardware, wu2019visionisp, yoshimura2023dynamicisp}. Although these methods can be regarded as RAW-aware detection approaches, they still heavily rely on ISP processing and do not achieve fully end-to-end RAW object detection. 
A straightforward alternative is to directly train existing sRGB-based detectors on RAW images. However, this usually leads to degraded performance~\cite{xu2023toward, li2024towards, cui2024raw}, mainly for two reasons. 
First, conventional detectors are designed for sRGB images and are not tailored to the unique characteristics of RAW data. 
Second, most detectors benefit from ImageNet-pretrained weights~\cite{deng2009imagenet}, while transferring sRGB-pretrained models to RAW inputs introduces a clear domain gap. 
To mitigate these issues, recent methods commonly introduce learnable ISPs that convert RAW data into sRGB-like images by mimicking human visual priors, such as denoising, gamma correction, and white balance~\cite{xu2023toward,cui2024raw,onzon2021neural}. 
Although effective, this RAW-to-sRGB paradigm optimizes RAW representation learning toward image reconstruction rather than the detection objective. 
As a result, it may introduce information bottlenecks and inherit human-centric priors that are not necessarily optimal for machine vision models. 

Instead of reconstructing visually plausible sRGB images, we argue that RAW images should be directly adapted to detection-oriented feature learning. 
From the perspective of detector architectures, most modern detectors consist of a backbone network for feature extraction~\cite{he2016deep,liu2021swin}, followed by necks, detection heads, or Transformer encoder-decoder modules for object localization and classification~\cite{ge2021yolox,zhu2020deformable}. 
These downstream components are largely designed for high-level semantic reasoning and should not be heavily modified merely to accommodate RAW inputs. 
Notably, within the backbone, the first convolutional layer directly interacts with image measurements. This observation motivates us to design a lightweight RAW feature adapter at the input stage, which addresses the unique characteristics of RAW data while preserving the general architecture of existing detectors.

\begin{figure}[t]
	\centering
	\begin{subfigure}[b]{0.70\linewidth}
		\includegraphics[width=0.49\linewidth]{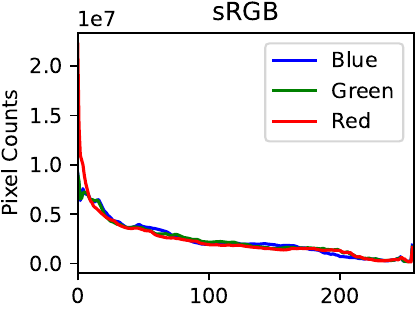}
		\includegraphics[width=0.49\linewidth]{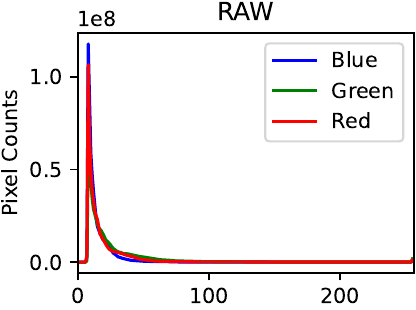}
		\caption{Histogram of pixel values}
	\end{subfigure}
	\hfill
	\begin{subfigure}[b]{0.28\linewidth}
		\centering
		\setlength{\tabcolsep}{3.5pt}
		\resizebox{1\linewidth}{!}{
			\begin{tabular}{lcc}
				\toprule
				& sRGB & RAW \\
				\midrule
				Red   & 1.001 & 0.788 \\
				Green & 1.032 & 1.257 \\
				Blue  & 0.967 & 0.955 \\
				\bottomrule
			\end{tabular}
		}
		\vspace{3mm}
		\caption{Normalized channel intensity}
	\end{subfigure}
	
	\vspace{0.5em}
	
	\begin{subfigure}[htbp]{\textwidth}
	\centering

	\begin{minipage}[c]{0.23\linewidth}
		\centering
		\includegraphics[width=\linewidth]{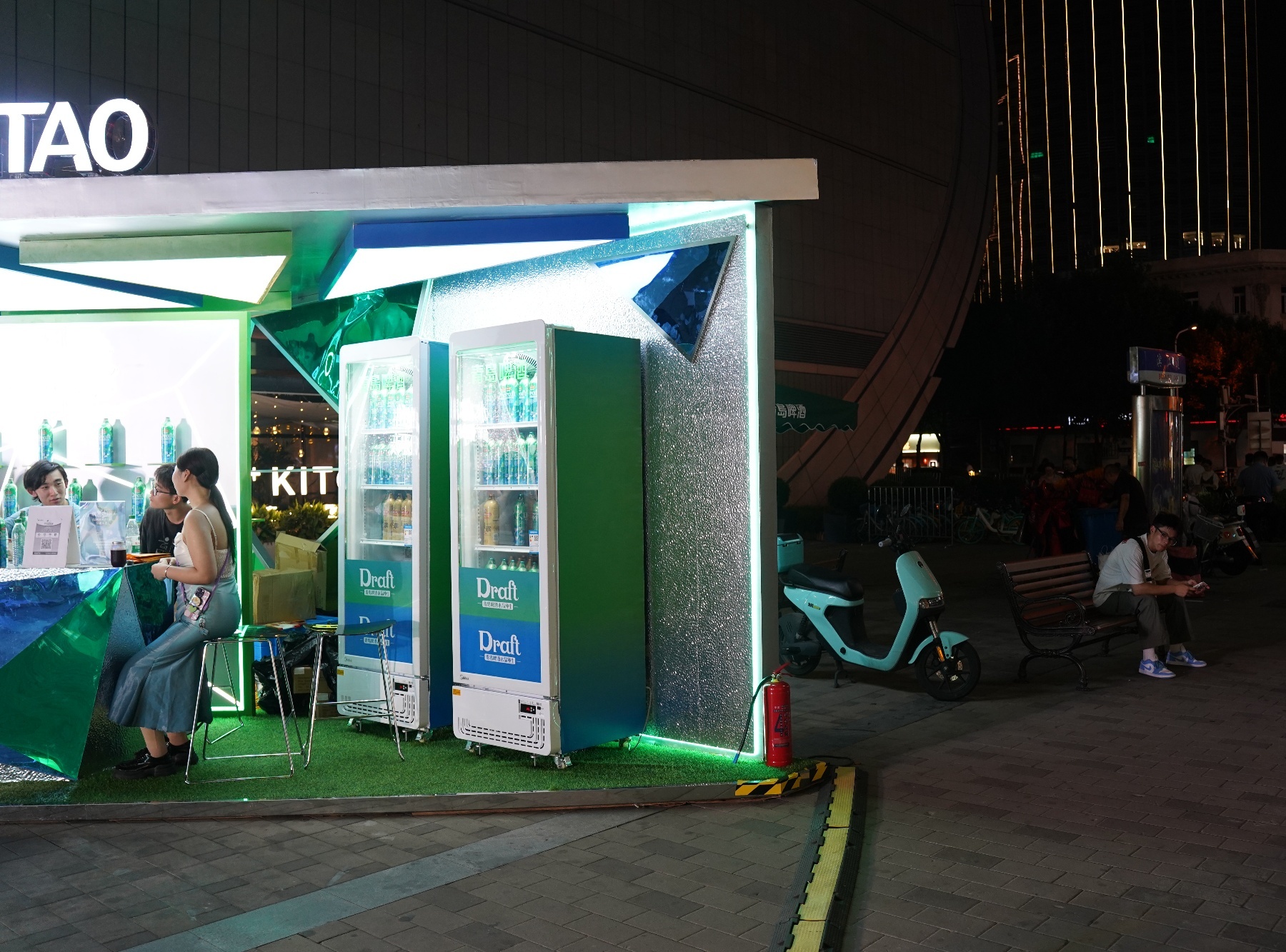}
		\par\small{sRGB}
	\end{minipage}
	\hfill
	\begin{minipage}[c]{0.23\linewidth}
		\centering
		\includegraphics[width=\linewidth]{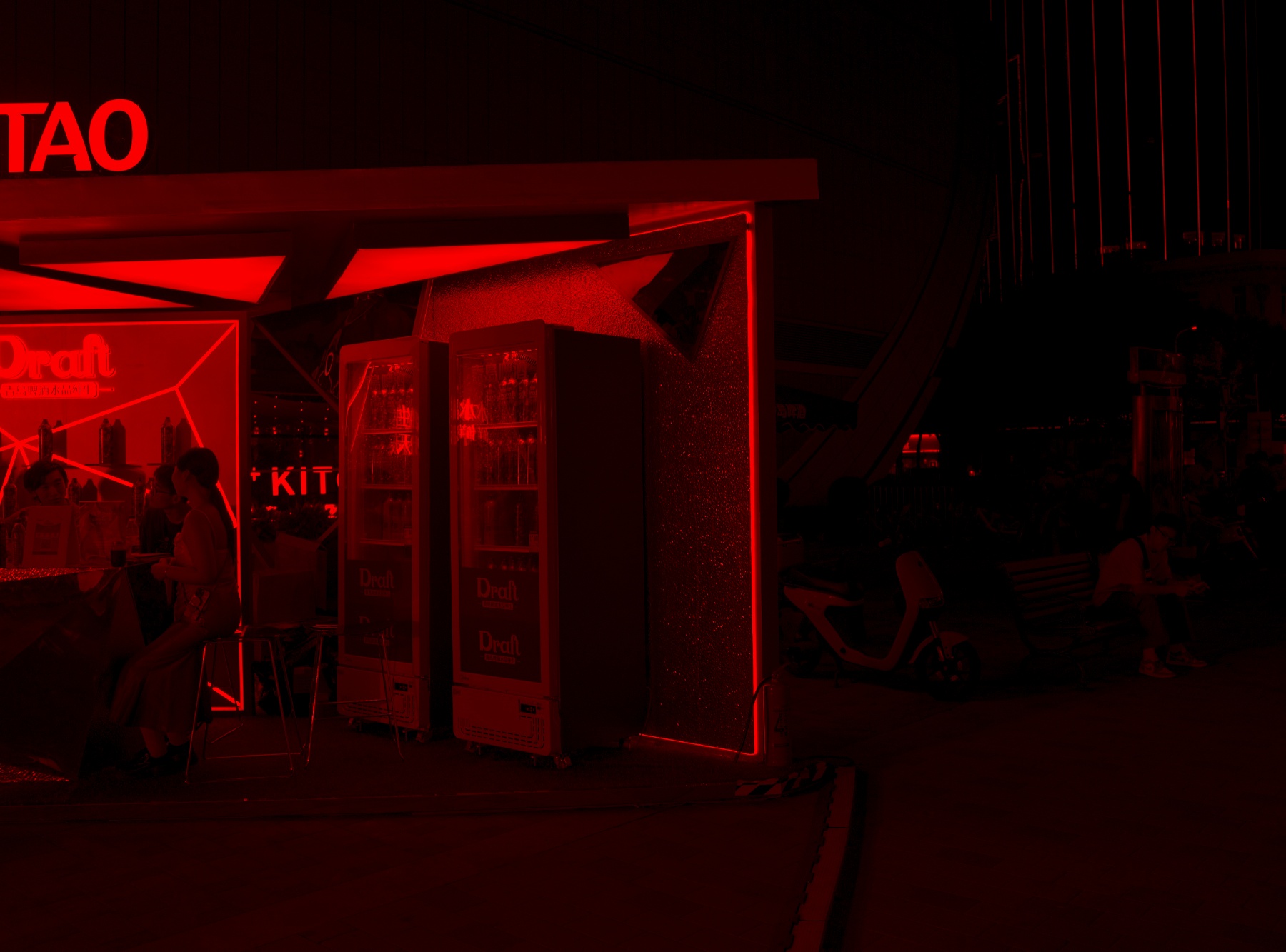}
		\par\small{RAW-Red}
	\end{minipage}
	\hfill
	\begin{minipage}[c]{0.23\linewidth}
		\centering
		\includegraphics[width=\linewidth]{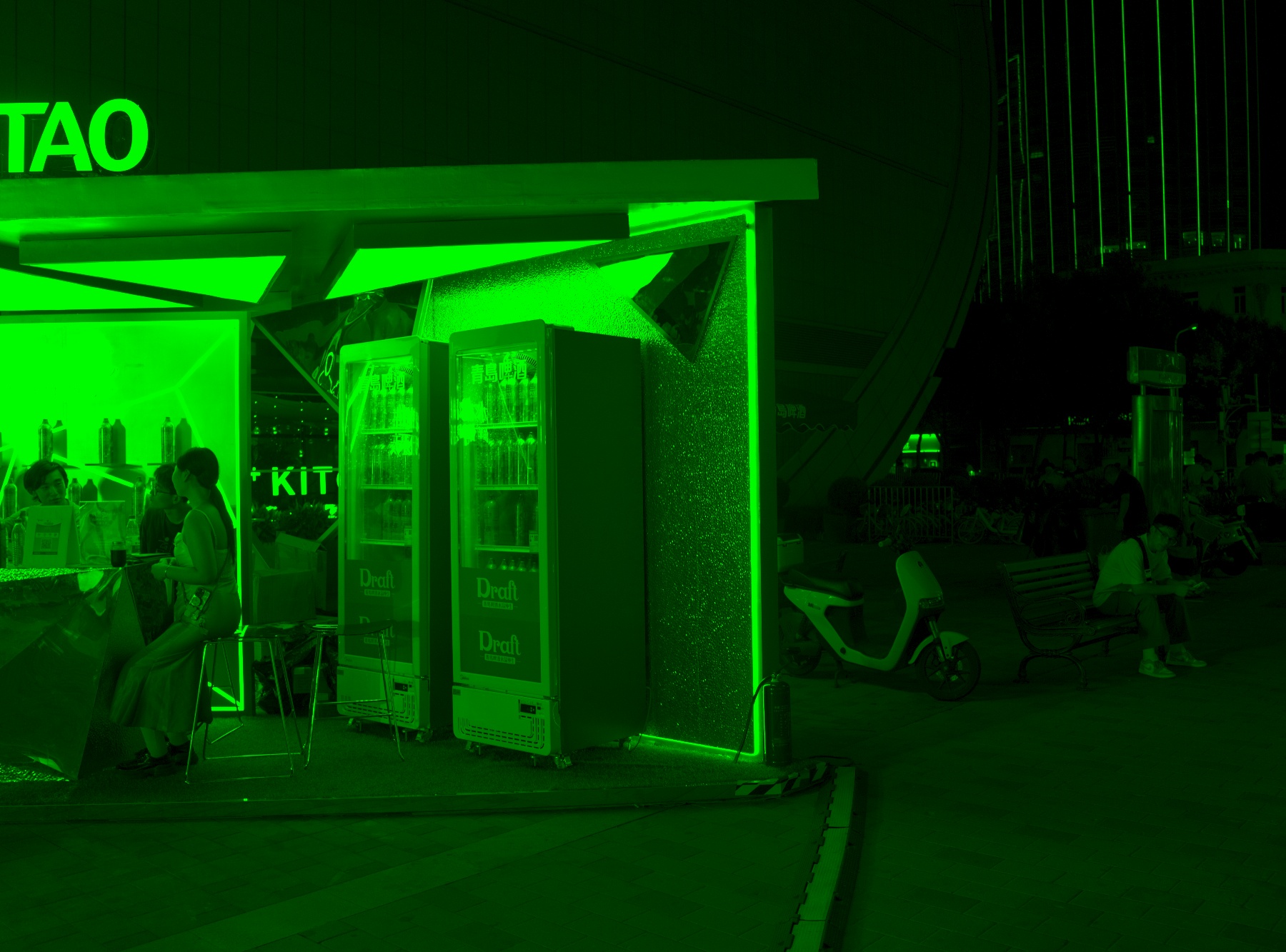}
		\par\small{RAW-Green}
	\end{minipage}
	\hfill
	\begin{minipage}[c]{0.23\linewidth}
		\centering
		\includegraphics[width=\linewidth]{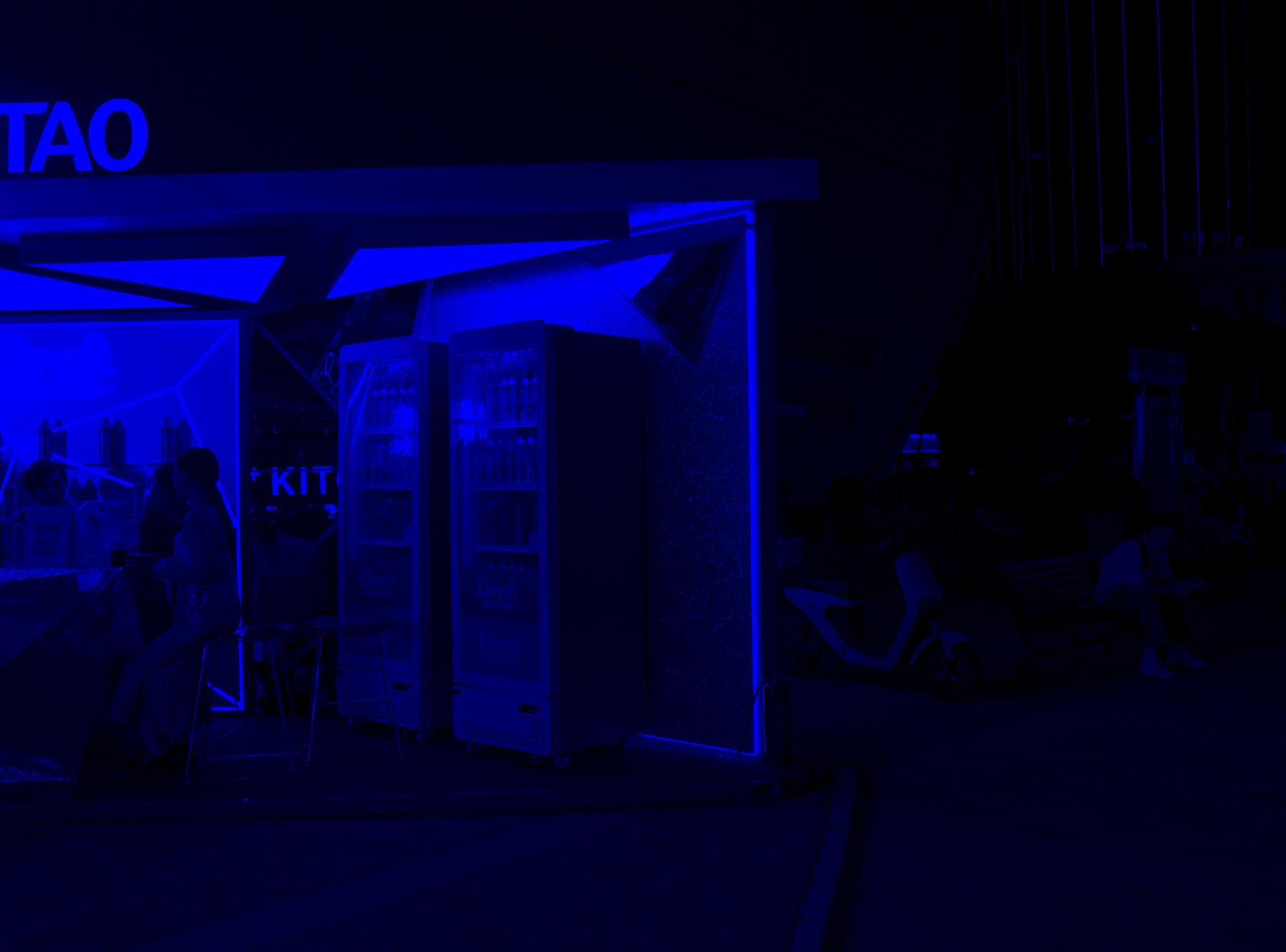}
		\par\small{RAW-Blue}
	\end{minipage}

	\caption{Visualization of different channels in RAW images}
    \end{subfigure}
	\vspace{-0.5em}
	\caption{Comparison of sRGB and RAW images. (\textcolor{red}{a}) RAW pixel values are concentrated within a small range. (\textcolor{red}{b}) RAW images exhibit significant channel differences. (\textcolor{red}{c}) In the example scene, the green channel is brighter, while the red channel is darker.}
	\label{fig:analyze}
	\vspace{-1em}
\end{figure}
We further analyze the inherent illumination characteristics of RAW images. As shown in Figure~\ref{fig:analyze}(a), the intensity distribution of RAW images is often highly concentrated within a narrow range, especially under low-light conditions. In contrast, sRGB images benefit from ISP operations such as gamma correction and tone mapping, which redistribute the dynamic range and produce a more balanced brightness distribution, thereby improving detection performance~\cite{li2024efficient}. However, the concentrated RAW intensity distribution does not indicate the absence of useful information. Instead, RAW pixels preserve a linear response to incident light intensity, making their brightness values closely related to the physical illumination of the scene. Therefore, the key challenge is how to effectively perceive and exploit spatial brightness variations in RAW images. To this end, we introduce an illumination mask to identify regions with different brightness levels and use it to guide feature extraction under varying lighting conditions.

In addition to spatial brightness variation, RAW images also exhibit distinct channel-dependent light sensitivity. As shown in Figure~\ref{fig:analyze}(b), the green channel typically has higher average intensity values, suggesting that it preserves more information in darker regions. In contrast, the red and blue channels are more responsive in brighter regions that are close to saturation. This phenomenon is further illustrated in Figure~\ref{fig:analyze}(c), where the red channel clearly delineates object boundaries in bright areas where the green channel tends to become saturated. Based on these observations, we propose an illumination-aware end-to-end RAW object detection method, termed ILLRAW. Unlike previous approaches that rely on learnable ISPs to bridge RAW and sRGB domains, ILLRAW directly integrates with existing detectors through a lightweight feature adapter. Specifically, we first introduce an illumination mask estimation module to capture spatial light intensity distributions. Then, a channel-sensitive embedding module is designed to extract features with different light sensitivities under the guidance of the green and red/blue channels. To effectively integrate these complementary representations, we propose an illumination attention fusion mechanism that adaptively aggregates features according to the estimated illumination mask. Moreover, re-parameterization techniques are employed to minimize inference overhead. The proposed method can effectively process RAW data with low computational cost and good compatibility across different detection architectures. Extensive experiments on three RAW object detection datasets demonstrate that ILLRAW achieves state-of-the-art performance, validating its effectiveness and generalizability.

In summary, our main contributions are as follows:
\begin{itemize}
	\item We propose ILLRAW, an illumination-aware RAW object detection framework that leverages spatial and channel-wise illumination differences to enhance semantic understanding of RAW images.
	
	\item We design an illumination estimation module to extract spatial illumination cues, and a channel-sensitive embedding module to decouple features with different light sensitivities under channel-specific guidance.
	
	\item We introduce an illumination attention fusion mechanism, which utilizes the estimated illumination mask to adaptively fuse features from regions with varying brightness levels.
	
	
\end{itemize}
\section{Related Work}
\label{sec:Related}
In this section, we first review the fundamentals of RAW image processing. Then, we summarize current research on both sRGB-based and RAW-based object detection.

\noindent\textbf{RAW-based Image Processing.} Modern camera imaging systems comprise three main components, namely the optical lens, the image sensor, and the image signal processor. When light hits the sensor, it is converted into electrical signals, then into a raw digital signal by an analog-to-digital converter (ADC). This raw signal is called a RAW image \cite{wei2020physics}. The RAW image is processed by the ISP and converted into an sRGB image suitable for human viewing. Traditional ISP designs are based on how humans perceive color and brightness \cite{karaimer2016software, nishimura2018automatic, ramanath2005color}. The ISP comprises several processing stages, each serving a specific purpose, although accurately performing these operations remains challenging. Since sRGB images are widely used, many image enhancement methods focus on this format \cite{dong2014learning, zhang2020deblurring, peng2018deep, zamir2022restormer, he2010single}. Recently, there is growing interest in using RAW images for tasks such as denoising \cite{brooks2019unprocessing, wei2020physics, jin2023dnf}, high dynamic range (HDR) imaging \cite{kronander2013unified, zou2023rawhdr, shu2024towards}, and general image enhancement \cite{chen2018learning, huang2022towards, cheng2023recaptured, conde2024bsraw}. RAW images have clear benefits in handling real-world problems, especially in HDR and low-light situations, but their potential for visual recognition tasks remains underexplored.

\noindent\textbf{sRGB-based Object Detection.} Object detection has greatly benefited from large-scale sRGB benchmarks \cite{everingham2015pascal, lin2014microsoft, shao2019objects365}. Early object detection methods primarily relied on hand-crafted features and sliding window techniques \cite{viola2001rapid, dalal2005histograms, felzenszwalb2009object}. Although these approaches have low computational cost, their limited generalization ability makes them less effective in complex scenarios. With the advent of deep learning, CNN-based detectors have achieved significant progress. These can be broadly divided into two-stage detectors \cite{girshick2015fast, ren2015faster, cai2018cascade, lin2017feature} and one-stage detectors \cite{redmon2016you, lin2017focal, redmon2018yolov3, duan2019centernet, ge2021yolox, wang2025yolov10}. More recently, transformer-based detectors \cite{zhu2020deformable, liu2021swin, zhang2022dino, zong2023detrs} have introduced global modeling capabilities and reformulated object detection as a set prediction problem, eliminating the need for anchors and non-maximum suppression (NMS). Despite the impressive performance of these methods, their robustness under real-world conditions remains a challenge \cite{hong2021crafting, loh2019getting, li2024towards}.

\noindent\textbf{RAW-based Object Detection.} RAW-based object detection holds promise for addressing challenging real-world scenarios. A straightforward approach is to train detectors directly on RAW images \cite{hong2021crafting, xu2023toward}. However, due to the lack of designs tailored specifically for RAW data, these methods often achieve limited performance. Some approaches focus on tuning non-differentiable ISP components \cite{mosleh2020hardware, tseng2019hyperparameter}, while others learn differentiable ISP networks \cite{cui2024raw, diamond2021dirty, wu2019visionisp, yoshimura2023dynamicisp, wang2025adaptiveisp} to bridge sRGB detectors with RAW inputs. For example, Xu et al. \cite{xu2023toward} explore image- and pixel-level information to adjust the dynamic range of RAW data, while Cui et al. \cite{cui2024raw} attempt to adapt sRGB-pretrained models to RAW images. These methods incorporate priors inspired by human visual perception, yet there is no evidence that such priors are optimal for machine vision tasks. Moreover, this decoupling strategy can cause information bottlenecks and may lead to inefficiencies. In contrast, we propose a universal RAW detection method that directly leverages the inherent characteristics of RAW images to improve performance.

\section{Method}
\label{sec:Method}
\begin{figure*}[tbp]
	\centering
	\includegraphics[width=1\linewidth]{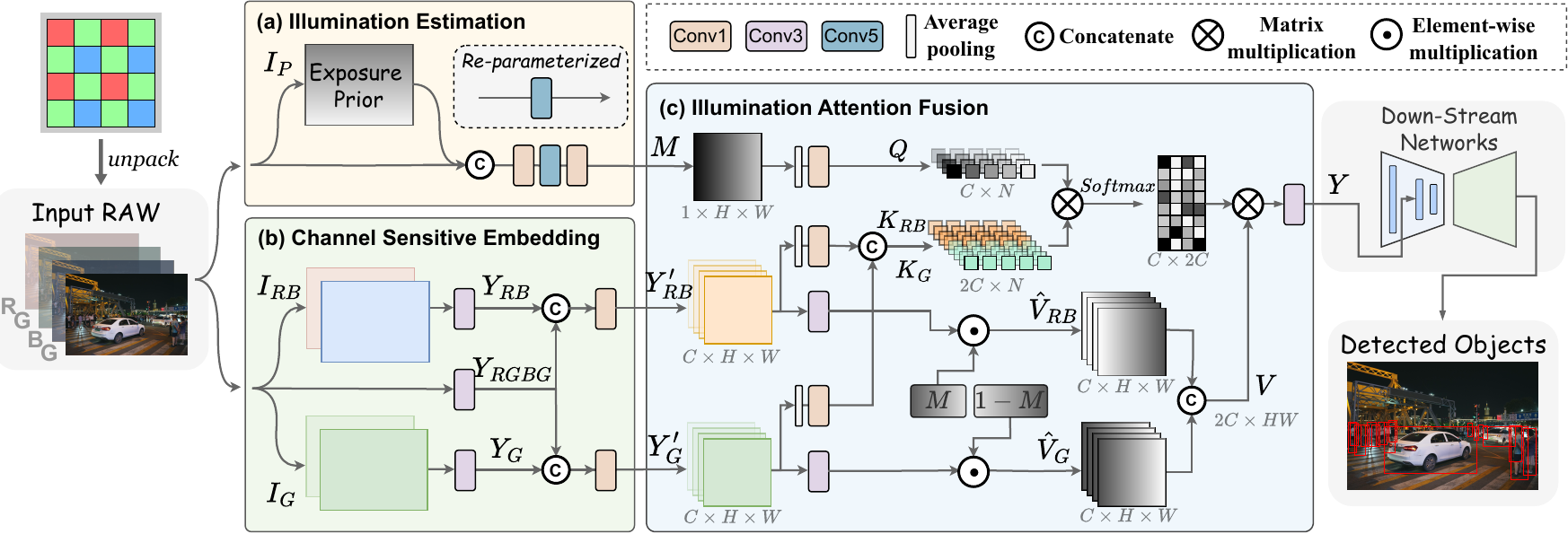}
	\caption{Overview of the proposed \textbf{ILLRAW}. It leverages illumination variations across the channel and spatial dimensions to effectively extract semantic information from RAW images.}
	\label{fig:framework}
	\vspace{-2em}
\end{figure*}

In this section, we first present the motivation behind our design. We then provide a detailed description of the proposed ILLRAW framework, followed by the training and inference strategies. The architecture is illustrated in Figure \ref{fig:framework}.

\subsection{Motivation}
\label{sec:motivation}


Existing RAW-based detectors commonly rely on learnable ISPs to convert RAW data into sRGB-like images by mimicking human visual priors, such as denoising, gamma correction, and white balance~\cite{xu2023toward,cui2024raw}. Although effective, this RAW-to-sRGB paradigm decouples RAW representation learning from the detection objective and may introduce information bottlenecks. Instead of reconstructing visually plausible sRGB images, we aim to directly adapt RAW inputs to detection-oriented feature learning. Since most modern detectors consist of a backbone for feature extraction~\cite{he2016deep,liu2021swin}, followed by task-specific necks, heads, or Transformer encoder-decoder modules for localization and classification~\cite{ge2021yolox,zhu2020deformable}, we preserve these downstream components and focus on the input stage, where the detector first interacts with image measurements.

Our adapter design is motivated by two RAW-specific properties. First, RAW pixels are linearly related to incident light intensity and exhibit strong spatial brightness variations, motivating an illumination mask estimation module. Second, RAW channels have different light sensitivities: the green channel and red/blue channels provide complementary cues under different illumination conditions, motivating a channel-sensitive embedding module. Finally, we introduce an illumination attention fusion module to adaptively aggregate these complementary features according to the estimated illumination mask.

\subsection{Architecture Design}

We propose a feature adaptation module that bridges the gap between sRGB-trained detectors and RAW images. This module replaces the detector’s initial layer and is jointly optimized with the rest of the network, enabling seamless integration with various detection architectures while maintaining strong performance.

Given an input RAW image $\mathbf{X}$, we first estimate an illumination mask using an illumination estimation module (IE). Next, a channel-sensitive embedding (CSE) module extracts features with varying sensitivity to bright and dark regions, guided by the red/blue and green channels, respectively. Finally, an illumination attention fusion (IAF) module selectively integrates the extracted features under the guidance of the estimated illumination mask. The overall architecture is illustrated in Figure \ref{fig:framework}.

\noindent \textbf{Illumination Estimation.}
RAW images contain abundant information in both bright and dark regions, which pose distinct challenges for object detectors. Specifically, dark regions often exhibit considerable sensor noise, while bright regions are prone to overexposure. To enable region-specific processing, we estimate an illumination mask that highlights the spatial distribution of brightness levels across the scene. Previous methods typically rely on Retinex theory~\cite{wei2018deep} to learn illumination estimators via deep networks. However, these approaches tend to introduce significant computational overhead. In contrast, we adopt a lightweight yet effective strategy by utilizing the mean value of the RAW image along the channel dimension as an illumination prior. This prior is concatenated with the original input and passed to a compact estimator $\mathcal{E}$ to produce the illumination mask.
\vspace{-0.5em}
\begin{equation}
	\mathbf{M} = \mathcal{E} \Big( \operatorname{Concat}\big( \mathbf{X}, \operatorname{Mean}_c(\mathbf{X}) \big) \Big),
\end{equation}
where $\mathcal{E}$ denotes the illumination estimator, $\mathbf{M}$ is the predicted illumination mask, and $\operatorname{Concat}$ and $\operatorname{Mean}_c$ represent the channel-wise concatenation and mean operations, respectively.

The architecture of the illumination estimator $\mathcal{E}$ is illustrated in Figure~\ref{fig:framework} (\textcolor{red}{a}). It first employs a $1 \times 1$ convolutional layer to fuse the concatenated input. To improve discrimination between bright and dark regions, a $5 \times 5$ convolutional layer is then applied to capture spatial context. Finally, another $1 \times 1$ convolutional layer is used to generate the illumination mask. Notably, we avoid any nonlinear activation functions in the design of $\mathcal{E}$, which enables the illumination prior and estimator $\mathcal{E}$ to be reparameterized into a single convolution during inference. More details on this reparameterization are provided in Section~\ref{sec:reparam}.

\noindent \textbf{Channel-Sensitive Embedding.}
Backbones designed for sRGB images, such as CNNs and Transformers, typically adopt a single convolutional layer to embed input images into feature space, implicitly treating all three color channels with equal importance. 
However, this assumption does not hold for RAW images, which exhibit substantial variations across different color channels. 
Notably, the green channel contains twice as many samples as the red or blue channels due to the Bayer pattern. 
Since the color channels share similar edge structures, modern ISPs often use the green channel to guide the spatial reconstruction of red and blue channels. 
However, this domain-specific prior has not been explicitly utilized in existing RAW-based object detection methods. 
Moreover, the three color channels demonstrate distinct sensitivities to illumination intensity. 
Motivated by these observations, we propose to learn independent embedding layers, each guided by either the green or the red/blue channels, thereby enhancing the model’s capability to recognize objects in regions with varying brightness levels.

The proposed channel-sensitive embedding module is illustrated in Figure~\ref{fig:framework} (\textcolor{red}{b}). It begins by unpacking the input RAW image into three branches. The basic branch includes all color channels and applies a $3 \times 3$ convolution to extract a general feature map $\mathbf{Y}_{RGBG}$. Two guidance branches are constructed, one for the green channel and another for the red and blue channels. Each branch passes through a $3 \times 3$ convolution to generate features $\mathbf{Y}_{G}$ and $\mathbf{Y}_{RB}$, respectively. Next, the features from the basic branch are concatenated with those from each guidance branch and fused through lightweight aggregation modules.
\begin{equation}
	\begin{aligned}
		\mathbf{Y}_{RB}' &= \mathcal{F}_{RB} \Big( \operatorname{Concat}(\mathbf{Y}_{RB}, \mathbf{Y}_{RGBG}) \Big) \\
		\mathbf{Y}_{G}'  &= \mathcal{F}_{G} \Big( \operatorname{Concat}(\mathbf{Y}_{G}, \mathbf{Y}_{RGBG}) \Big),
	\end{aligned}
\end{equation}
where $\mathcal{F}_{RB}$ and $\mathcal{F}_{G}$ denote the aggregation modules for the red/blue and green branches, respectively. $\mathbf{Y}_{RB}'$ and $\mathbf{Y}_{G}'$ are the resulting channel-sensitive embedded features. For simplicity and efficiency, each aggregation module is implemented as a $1 \times 1$ convolution, enabling reparameterization before inference. This design allows the module to be equivalently simplified into two independent $3 \times 3$ convolutions at test time, without additional computational cost.

\noindent \textbf{Illumination Attention Fusion.}
After obtaining the illumination mask $\mathbf{M}$ and the embedded features $\mathbf{Y}_{RB}'$ and $\mathbf{Y}_{G}'$, we perform feature fusion under illumination guidance. Conventional addition or concatenation applies fixed fusion rules and cannot adapt to spatial brightness variations. To address this limitation, the proposed IAF module uses $\mathbf{M}$ to guide dynamic and global feature interaction.

As illustrated in Figure~\ref{fig:framework} (\textcolor{red}{c}), IAF is built upon cross-attention. Specifically, the illumination mask is projected into the query $\mathbf{Q}$, which provides spatial intensity guidance, while the two embedded features are separately projected into key--value pairs $(\mathbf{K}_{RB},\mathbf{V}_{RB})$ and $(\mathbf{K}_{G},\mathbf{V}_{G})$. The keys encode global channel relationships for attention estimation, whereas the values retain the branch-specific features to be fused. To reduce computational complexity, adaptive average pooling is applied only to the query and key branches. Following the channel-first layout used in our implementation and omitting the batch dimension, let the pooled spatial size be $h\times w$, with $N=hw$, and let $H\times W$ denote the original resolution. For each attention head, $\mathbf{Q},\mathbf{K}_{RB},\mathbf{K}_{G}\in\mathbb{R}^{d_k\times N}$, where $d_k$ is the channel dimension of each head. The two keys are concatenated along the channel dimension as $\mathbf{K}\in\mathbb{R}^{2d_k\times N}$. Meanwhile, $\mathbf{V}_{RB},\mathbf{V}_{G}\in\mathbb{R}^{d_k\times HW}$ retain the original spatial resolution. They are modulated by $\mathbf{M}$ and $1-\mathbf{M}$, respectively, and concatenated along the channel dimension as $\mathbf{V}\in\mathbb{R}^{2d_k\times HW}$. The fused feature is computed as
\begin{equation}
	\mathbf{Y} = \operatorname{softmax} \!\Bigg( \frac{\mathbf{Q}\mathbf{K}^{\top}}{\sqrt{d_k}} \Bigg) \mathbf{V},
\end{equation}
where the softmax is applied along the last dimension of $\mathbf{Q}\mathbf{K}^{\top}$. Accordingly, $\mathbf{Q}\mathbf{K}^{\top}\in\mathbb{R}^{d_k\times 2d_k}$ captures global interactions between the query channels and the concatenated RB/G key channels, while $\mathbf{Y}\in\mathbb{R}^{d_k\times HW}$ preserves the original spatial resolution. Finally, the outputs of all heads are concatenated and projected by a $3\times3$ convolution.

\subsection{Training and Inference}

\textbf{Training details.}
The proposed method is jointly optimized with the detector and can be integrated into representative CNN- and Transformer-based detection frameworks by replacing the input embedding layer. It utilizes the default loss functions of the underlying detector without requiring additional modifications. Although detector performance benefits significantly from ImageNet-pretrained weights, our method modifies the original network architecture, which may limit the full utilization of pretrained features and exacerbate the domain gap. Nevertheless, we observe that ImageNet-pretrained weights still provide substantial performance gains for our approach. A detailed analysis is presented in Section \ref{sec:ablation}.

\noindent \textbf{Inference.}
\label{sec:reparam}
To reduce computational overhead during inference, we employ a reparameterization strategy for both the illumination estimation (IE) and channel-sensitive embedding (CSE) modules. These modules are designed exclusively with convolutional and concatenation operations, enabling straightforward fusion. The channel-wise mean operation can be represented as a $1 \times 1$ convolution with kernel size $1 \times C \times 1 \times 1$, where all kernel weights are set to $1/C$. For the green and red/blue guidance branches, missing color channels are aligned with the basic branch by introducing zero-valued convolutional kernels, allowing the branches to be merged seamlessly. Consequently, at inference time, the IE module requires only a single $5 \times 5$ convolution, and the CSE module involves two $3 \times 3$ convolutions, thereby substantially reducing computational cost.

\section{Experiments}
\label{sec:Experiments}
In this section, we first describe the experimental setup. Next, we present detailed results of our experiments. Finally, we perform ablation studies to analyze the contributions of key components.

\subsection{Experimental Setup}

\noindent \textbf{Dataset.} We conduct experiments on three RAW datasets, namely AODRaw \cite{li2024towards}, PascalRAW \cite{omid2014pascalraw}, and LOD \cite{hong2021crafting}. The AODRaw dataset comprises 7,785 real RAW images annotated across four categories, covering diverse challenging conditions such as daylight, low-light, rain, and fog. We adhere to the official data splits provided. PascalRAW is a widely used RAW object detection benchmark. Following the evaluation protocol in \cite{cui2024raw}, we assess performance under three conditions, namely normal, overexposure, and dark. Note that the latter two conditions involve synthetically generated images based on \cite{cui2021multitask}. The LOD dataset focuses on real-world low-light object detection and contains 2,230 RAW images with annotations for eight categories. Following \cite{cui2024raw}, we use 1,800 images for training and 430 for testing.

\noindent \textbf{Comparison Methods.} To evaluate the effectiveness of the proposed method, we compare it against various open-source ISP pipelines and RAW-based object detection approaches. Karaimer \textit{et al.} \cite{karaimer2016software} introduce a manually designed traditional ISP pipeline. InvISP \cite{xing2021invertible} represents a network-based ISP method. RAOD \cite{xu2023toward} and RAW-Adapter \cite{cui2024raw} are two state-of-the-art RAW detection methods. For a comprehensive evaluation, experiments are conducted using multiple detectors. On the AODRaw dataset, we employ Cascade-RCNN \cite{cai2018cascade}, a two-stage CNN-based detector, and Deformable DETR \cite{zhu2020deformable}, a Transformer-based detector. For the PascalRAW and LOD datasets, we adopt RetinaNet \cite{lin2017focal}, a one-stage CNN-based detector. For methods reproduced in our implementation, we use the same data augmentation protocols and training configurations. Results taken directly from previous studies are explicitly marked with * in the corresponding tables.

\noindent \textbf{Implementation Details.} We implement all methods using the open-source object detection toolbox MMDetection \cite{chen2019mmdetection}. For the AODRaw dataset, Cascade-RCNN is trained for 12 epochs, while Deformable DETR is trained for 25 epochs. For the PascalRAW and LOD datasets, the detectors are trained for 50 and 35 epochs, respectively. A batch size of 4 is used for all detectors. Data augmentation includes random horizontal flipping and random resizing. All models reproduced in our implementation are initialized with ImageNet-1K-pretrained weights by default. Experiments are conducted on NVIDIA RTX 3090 GPUs.

\begin{table}[tbp]
	\centering
    \caption{Comparison results on the AODRaw dataset. We use mAP and AP under different weather conditions to evaluate the detection performance. Results marked with * correspond to the values reported in~\cite{li2024towards}. Here, C-RCNN and D-DETR denote Cascade-RCNN and Deformable DETR. Params and FPS denote the number of model parameters in millions and the inference speed measured on one NVIDIA RTX 3090 GPU, respectively.}
	\label{tab:aodraw}
	\scriptsize
	\setlength{\tabcolsep}{2.2pt}
	\renewcommand{\arraystretch}{0.95}
	\resizebox{\linewidth}{!}{
	\begin{tabular}{ll|ccc|cccc|cc}
		\toprule
		Detector & Method 
		& mAP & AP$_{50}$ & AP$_{75}$ 
		& Normal & Low & Rain & Fog 
		& Params & FPS \\ 
		\midrule
		C-RCNN & sRGB  
		& 18.4 & 29.5 & 19.5 & 21.0 & 16.1 & 15.7 & 14.4 & 60.96 & 40.8 \\
		C-RCNN & RAW*  
		& 18.1 & 30.9 & 18.3 & 20.6 & 16.7 & 17.3 & 14.6 & 60.96 & 40.8 \\
		C-RCNN & Karaimer \textit{et al.} 
		& 18.3 & 29.7 & 19.6 & 22.1 & 16.1 & 16.1 & 13.8 & 60.96 & 40.8 \\
		C-RCNN & InvISP  
		& 17.4 & 28.7 & 18.3 & 21.3 & 15.6 & 13.1 & 13.0 & 61.10 & 30.7 \\
		C-RCNN & RAOD 
		& 21.1 & 33.6 & 22.2 & 23.8 & 19.3 & 19.9 & 17.4 & 61.03 & 35.6 \\
		C-RCNN & RAW-Adapter* 
		& 19.9 & 33.2 & 20.1 & 22.3 & 18.1 & 20.8 & 16.9 & 61.11 & 36.2 \\
		\rowcolor{gray!20} 
		C-RCNN & \textbf{ILLRAW}
		& \textbf{23.3} & \textbf{36.0} & \textbf{24.8} 
		& \textbf{26.7} & \textbf{20.9} & \textbf{23.3} & \textbf{18.4} 
		& 61.10 & 38.1 \\
		\midrule
		D-DETR & sRGB* 
		& 16.6 & 31.9 & 15.6 & 18.3 & 15.2 & 16.4 & 13.1 & 40.11 & 43.1 \\
		D-DETR & RAW 
		& 20.5 & \textbf{36.9} & 20.5 & 23.9 & 18.5 & 19.6 & 16.2 & 40.11 & 43.1 \\
		D-DETR & Karaimer \textit{et al.} 
		& 20.4 & \textbf{36.9} & 20.4 & 23.5 & 18.8 & 18.1 & 16.5 & 40.11 & 43.1 \\
		D-DETR & InvISP 
		& 19.8 & 35.3 & 19.3 & 23.6 & 18.0 & 17.8 & 14.9 & 40.25 & 37.1 \\
		D-DETR & RAOD
		& 17.2 & 30.6 & 17.1 & 19.6 & 15.6 & 14.5 & 12.1 & 40.19 & 41.2 \\
		D-DETR & RAW-Adapter
		& 18.0 & 32.7 & 17.4 & 20.8 & 17.0 & 15.6 & 13.0 & 40.68 & 39.3 \\
		\rowcolor{gray!20} 
		D-DETR & \textbf{ILLRAW}
		& \textbf{22.3} & 36.6 & \textbf{23.5} 
		& \textbf{25.4} & \textbf{20.1} & \textbf{22.4} & \textbf{18.6} 
		& 40.26 & 42.3 \\
		\bottomrule
	\end{tabular}
	}
	\vspace{-2em}
\end{table}

\subsection{Experimental Results}
\label{sec:exp_result}

We perform comprehensive quantitative and qualitative evaluations to compare the performance of all methods on three benchmark datasets.

\noindent \textbf{Quantitative Comparison.} 
Table \ref{tab:aodraw} presents the results on the AODRaw dataset. The proposed method achieves highly competitive results across all evaluated detectors. Specifically, with the Cascade-RCNN detector, ILLRAW outperforms the strongest baseline RAOD by 2.2 mAP (improving from 21.1 to 23.3). Training directly on RAW data yields an mAP of 18.1. This value is 0.3 lower than the mAP of 18.4 obtained by training on sRGB images. The open-source ISP methods achieve moderate accuracy on Cascade-RCNN and do not surpass the sRGB-based detector. With Deformable DETR, direct RAW training and the Karaimer pipeline achieve 20.5 and 20.4 mAP, respectively, both outperforming the sRGB baseline of 16.6 mAP. This architecture-dependent result suggests that the relative effectiveness of RAW processing strategies varies with the underlying detector. Regarding RAW-specific detection methods, RAOD and RAW-Adapter achieve improvements of 3.0 and 1.8 mAP, respectively, over direct RAW training on Cascade-RCNN, demonstrating meaningful gains. With Deformable DETR, however, they achieve 17.2 and 18.0 mAP and therefore do not improve upon direct RAW training. In contrast, ILLRAW reaches 22.3 mAP, exceeding direct RAW training by 1.8 mAP. Analyzing performance under different weather conditions, our method maintains strong accuracy across all scenarios. On Cascade-RCNN, ILLRAW achieves the highest result in normal lighting conditions (26.7 AP), significantly outperforming other methods. Although performance drops under foggy conditions (18.4 AP), it still leads the comparison. Similar trends are observed with Deformable DETR, highlighting the robustness and adaptability of our approach to diverse environmental conditions. In terms of model complexity and computational efficiency, all methods introduce only marginal increases in parameter count compared to baseline detectors. However, inference speeds vary. InvISP, due to its reliance on a heavy open-source ISP pipeline, exhibits the lowest FPS. RAOD and RAW-Adapter also incur noticeable frame rate reductions. Our method yields the smallest decrease in inference speed, with a drop of only 2.7 FPS on Cascade-RCNN and 0.8 FPS on Deformable DETR, demonstrating an excellent trade-off between accuracy and efficiency.

\begin{table}[t]
	\centering
    \caption{Comparison results on the LOD dataset. The detector is RetinaNet~\cite{lin2017focal} with a ResNet-50 backbone. We report the AP for each category and the overall mAP. Results marked with * correspond to the values reported in~\cite{li2024towards}.}
	\label{tab:lod}
	\scriptsize
	\setlength{\tabcolsep}{2.2pt}
	\renewcommand{\arraystretch}{0.95}
	\resizebox{\linewidth}{!}{
	\begin{tabular}{l|cccccccc|c}
		\toprule
		Method 
		& Bicycle & Car & Motor. & Chair & Dining & Bottle & TV & Bus & Total \\
		\midrule
		sRGB* 
		& 67.6 & 89.1 & \textbf{59.9} & 77.3 & 39.8 & 34.8 & 50.9 & 55.9 & 59.4 \\
		RAW* 
		& 71.9 & 89.3 & 57.7 & 74.6 & 37.1 & 31.0 & 43.6 & 62.4 & 58.5 \\
		Karaimer \textit{et al.}*  
		& 66.9 & 87.3 & 57.6 & 71.3 & 30.2 & 29.7 & 35.8 & 56.0 & 54.4 \\
		InvISP* 
		& 72.6 & 89.2 & \textbf{59.9} & 76.8 & 38.9 & 31.9 & 31.4 & 56.6 & 57.1 \\
		RAOD 
		& 71.0 & 89.9 & 28.2 & \textbf{83.4} & 32.4 & 31.2 & 34.8 & 63.4 & 54.2 \\
		RAW-Adapter* 
		& \textbf{76.2} & 89.7 & 58.0 & 77.4 & \textbf{40.6} & \textbf{43.2} & 46.0 & 65.9 & 62.1 \\
		\rowcolor{gray!20}  
		\textbf{ILLRAW} 
		& 73.6 & \textbf{90.1} & 59.2 & 76.8 & 38.0 & 30.6 & \textbf{63.3} & \textbf{66.9} & \textbf{62.3} \\
		\bottomrule
	\end{tabular}
	}
	\vspace{-2em}
\end{table}

\begin{table}
	\centering
    \caption{Comparison results on the PascalRAW dataset. The detector is RetinaNet \cite{lin2017focal} with a ResNet-18 backbone. We present AP under three different illumination conditions. Results marked with * correspond to the values reported in \cite{cui2024raw}.}
	\label{tab:pascalraw}
		\begin{tabular}{l|ccc}
			\toprule
			Method & Normal & Overexposure & Dark \\
			\midrule
			sRGB* & 88.3 & - & - \\
			RAW* & 87.7 & 87.7 & 80.3 \\
			Karaimer \textit{et al.}* & 88.1 & 85.6 & 78.8 \\
			InvISP* & 85.4 & 86.6 & 70.9 \\
			RAOD & 88.1 & 88.1 & 83.9\\
			RAW-Adapter* & \textbf{88.7} & \textbf{88.7} & 82.5 \\
			\rowcolor{gray!20}  \textbf{ILLRAW (Ours)} & \textbf{88.7} & 88.5 & \textbf{84.7} \\
			\bottomrule
		\end{tabular}
	\vspace{-1em}
\end{table}
We evaluate the per-class performance on the LOD dataset using RetinaNet with a ResNet-50 backbone. As shown in Table \ref{tab:lod}, ILLRAW achieves the highest overall accuracy and outperforms RAW-Adapter by 0.2 mAP. Across categories, it achieves the best results in the car, TV monitor, and bus classes, while maintaining competitive performance in the remaining classes. In the TV monitor category, ILLRAW surpasses the strongest baseline by 12.4 AP. We also observe that open-source ISP methods fail to outperform even the simple RAW baseline.

Table~\ref{tab:pascalraw} reports the results on PascalRAW using RetinaNet with a ResNet-18 backbone. Under normal exposure, ILLRAW achieves 88.7 AP, tying RAW-Adapter for the best result. Direct RAW training obtains 87.7 AP, which is 0.6 AP lower than the sRGB baseline. Under overexposure, ILLRAW reaches 88.5 AP, 0.2 AP below RAW-Adapter and 0.4 AP above RAOD. Under dark exposure, ILLRAW achieves the best result of 84.7 AP, outperforming RAOD and RAW-Adapter by 0.8 and 2.2 AP, respectively. From overexposure to dark exposure, ILLRAW decreases by 3.8 AP, the smallest drop among the methods evaluated under both conditions. These results indicate that ILLRAW is particularly robust under dark exposure while remaining competitive under normal and overexposed conditions.

\begin{figure*}[htbp]
	\centering
	\begin{subfigure}[b]{0.24\linewidth}
		\includegraphics[width=\linewidth]{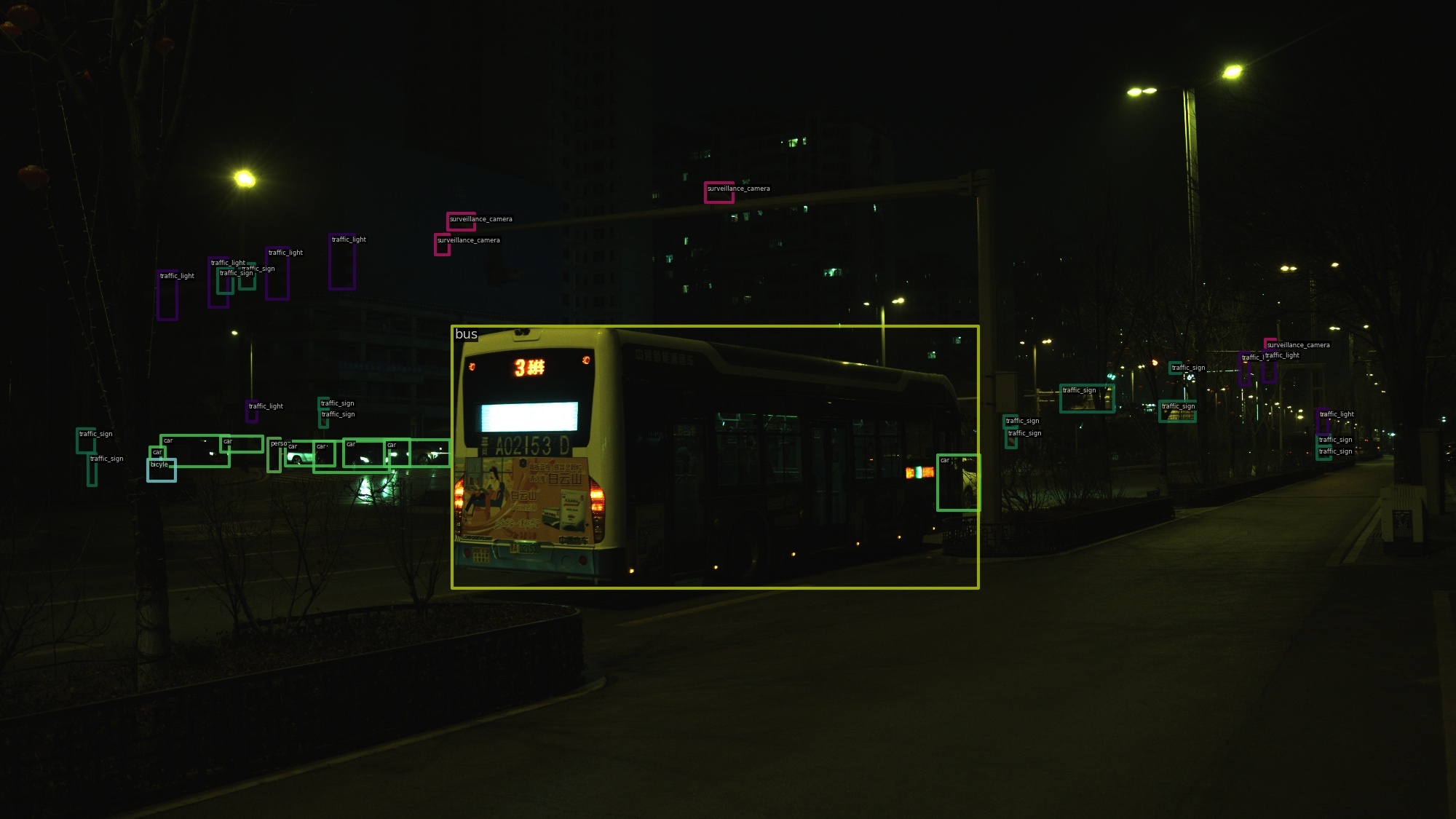}
		\caption{GT}
	\end{subfigure}
	\begin{subfigure}[b]{0.24\linewidth}
		\includegraphics[width=\linewidth]{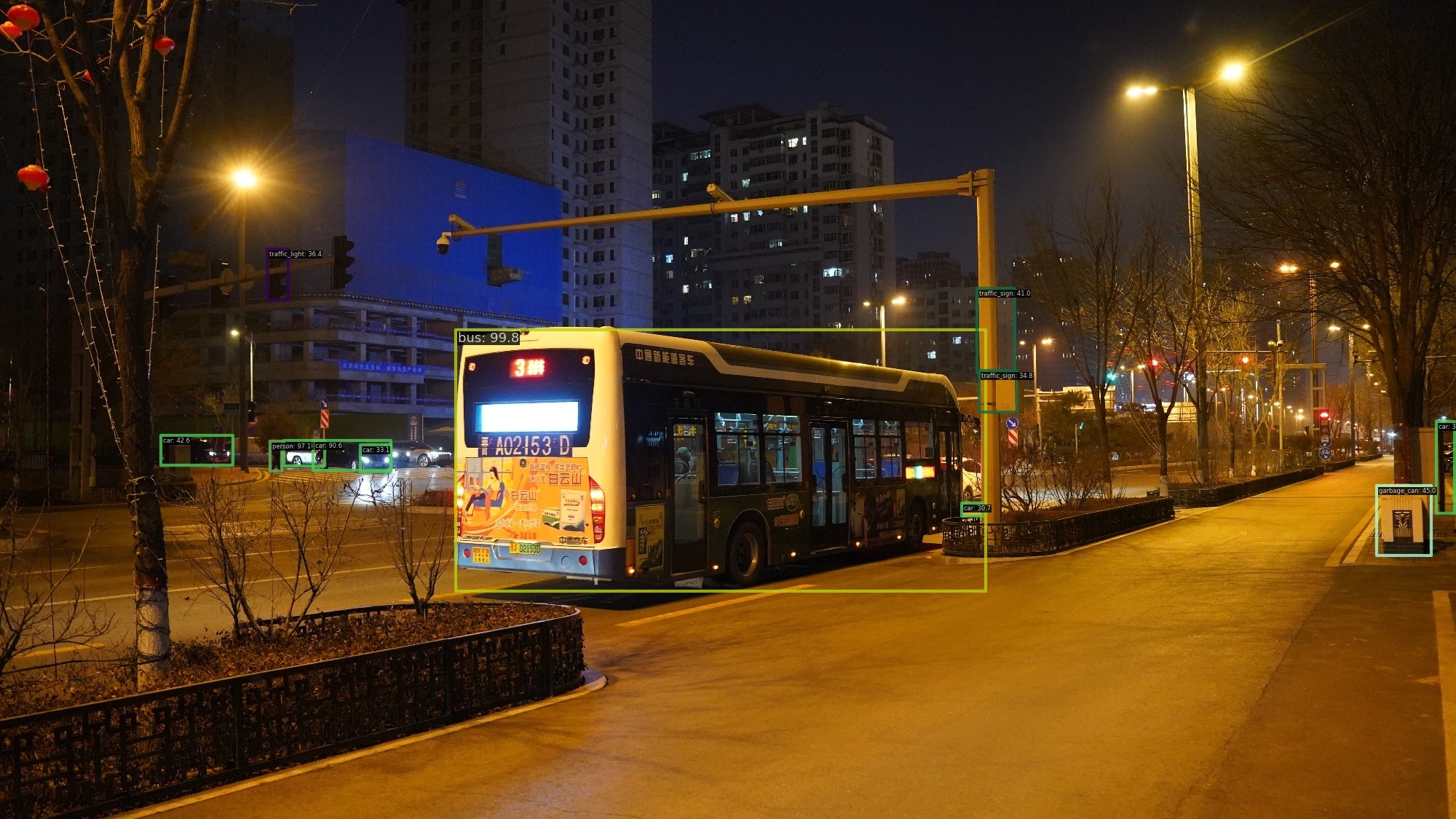}
		\caption{sRGB}
	\end{subfigure}
	\begin{subfigure}[b]{0.24\linewidth}
		\includegraphics[width=\linewidth]{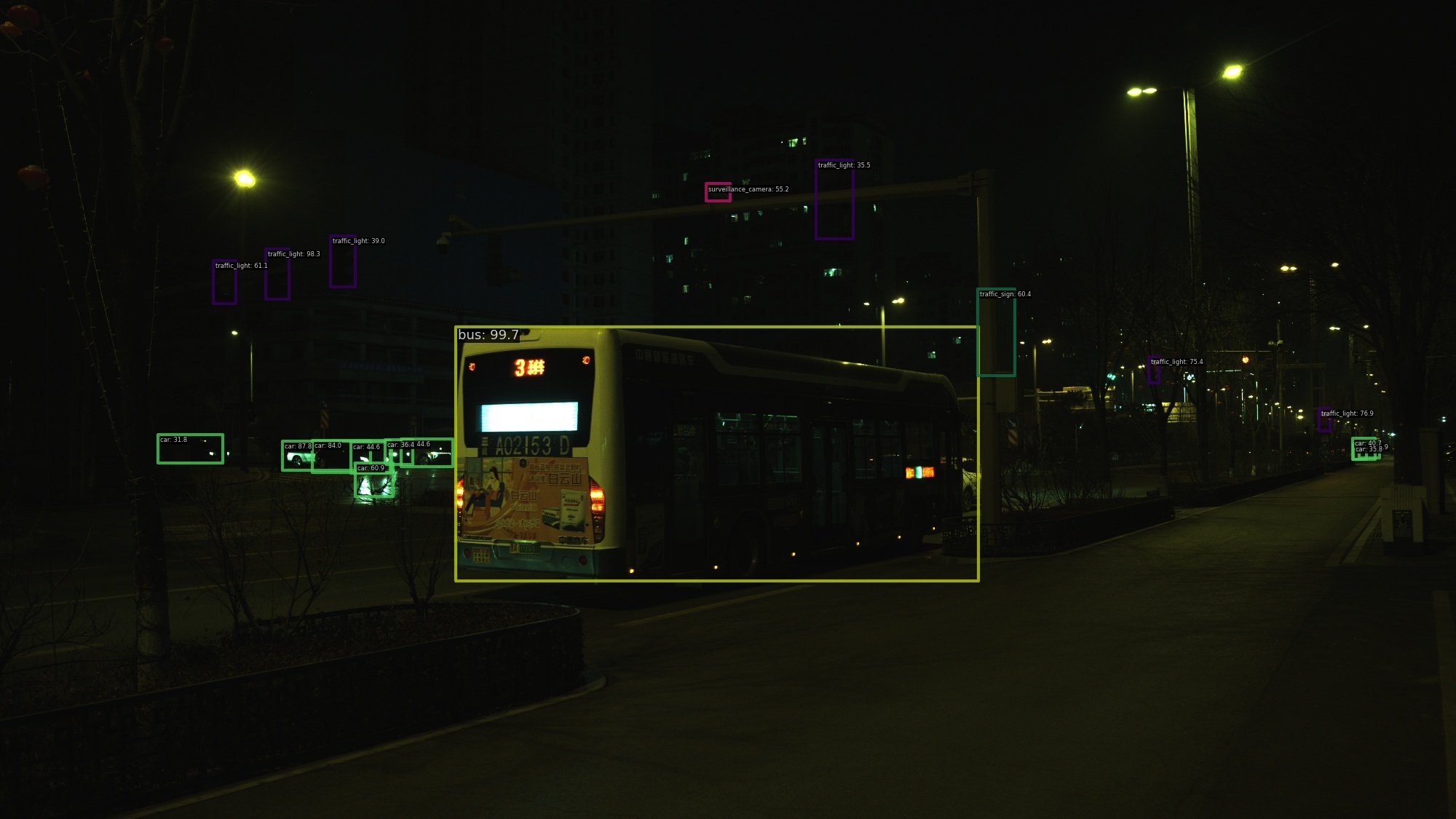}
		\caption{RAW}
	\end{subfigure}
	\begin{subfigure}[b]{0.24\linewidth}
		\includegraphics[width=\linewidth]{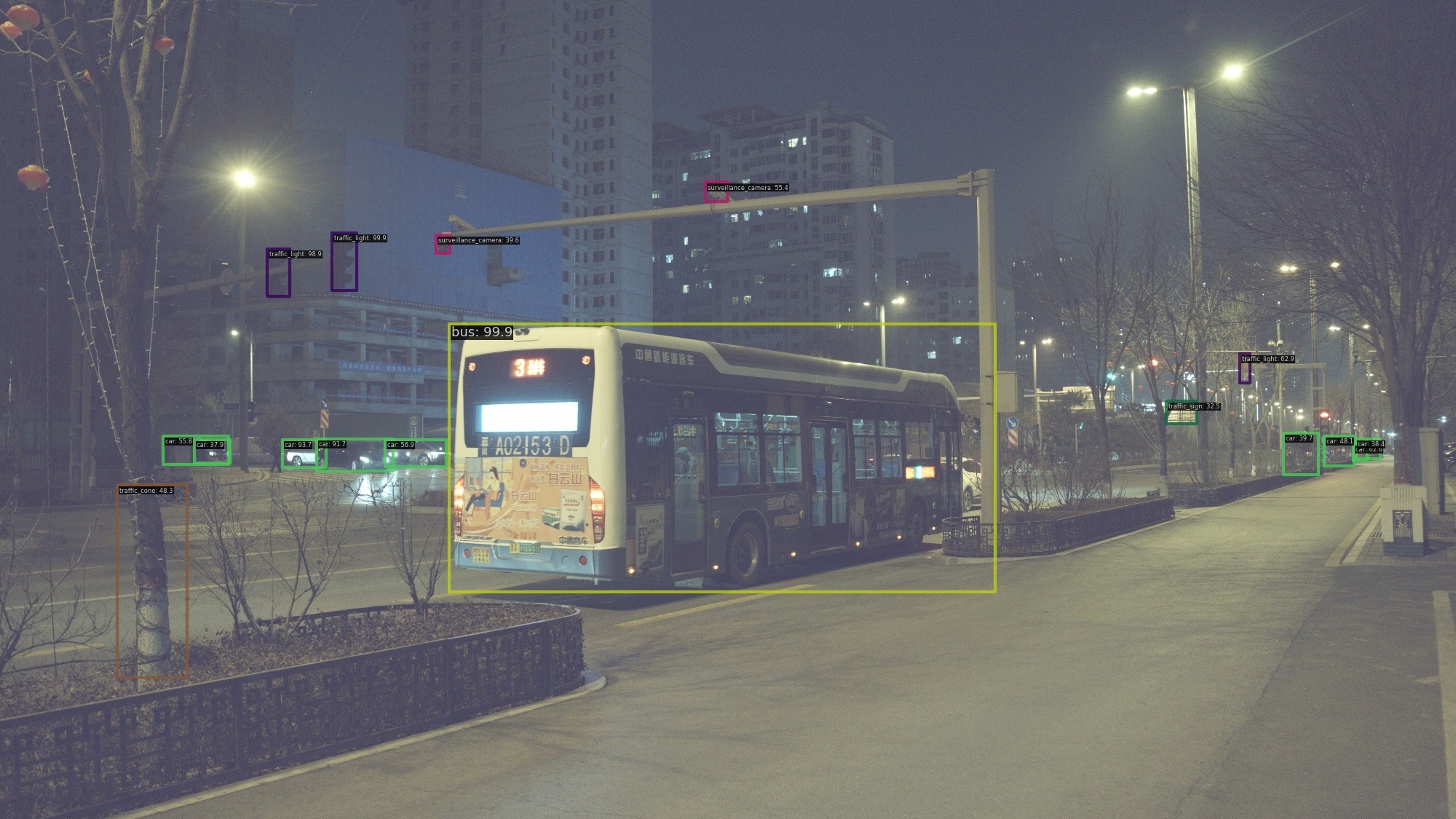}
		\caption{Karaimer \textit{et al.}}
	\end{subfigure} \\
	\begin{subfigure}[b]{0.24\linewidth}
		\includegraphics[width=\linewidth]{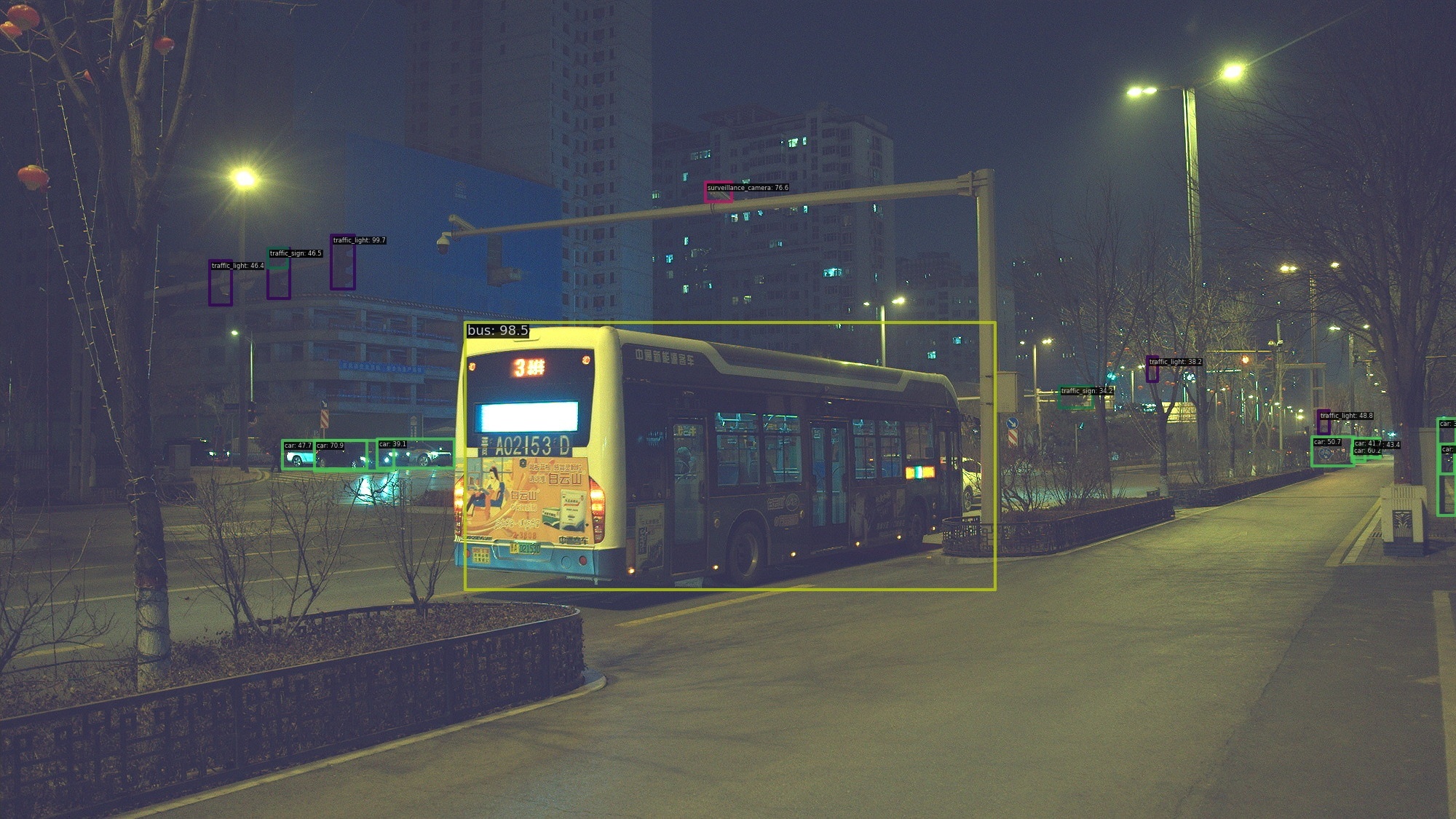}
		\caption{InvISP}
	\end{subfigure}
	\begin{subfigure}[b]{0.24\linewidth}
		\includegraphics[width=\linewidth]{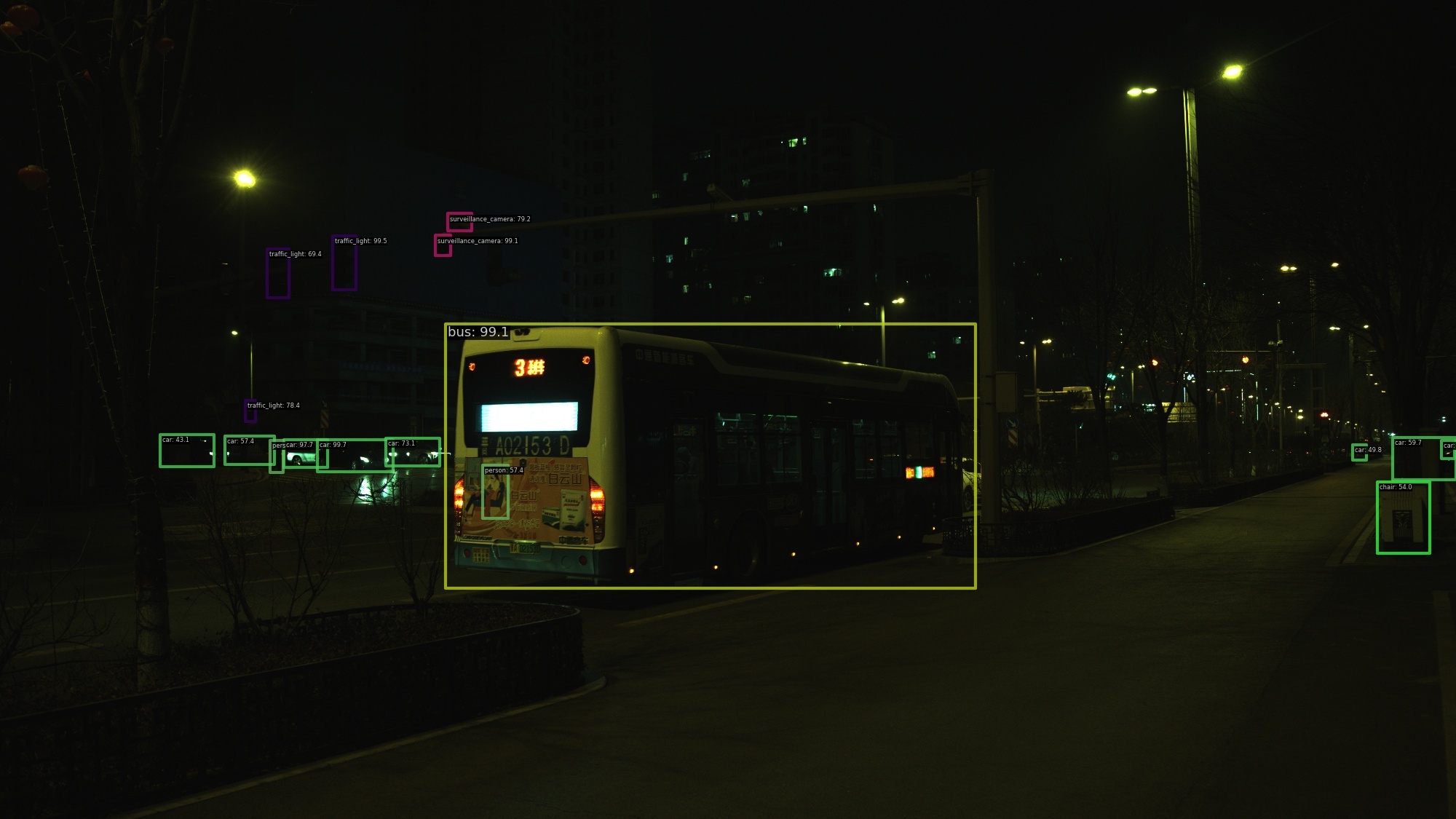}
		\caption{RAOD}
	\end{subfigure}
	\begin{subfigure}[b]{0.24\linewidth}
		\includegraphics[width=\linewidth]{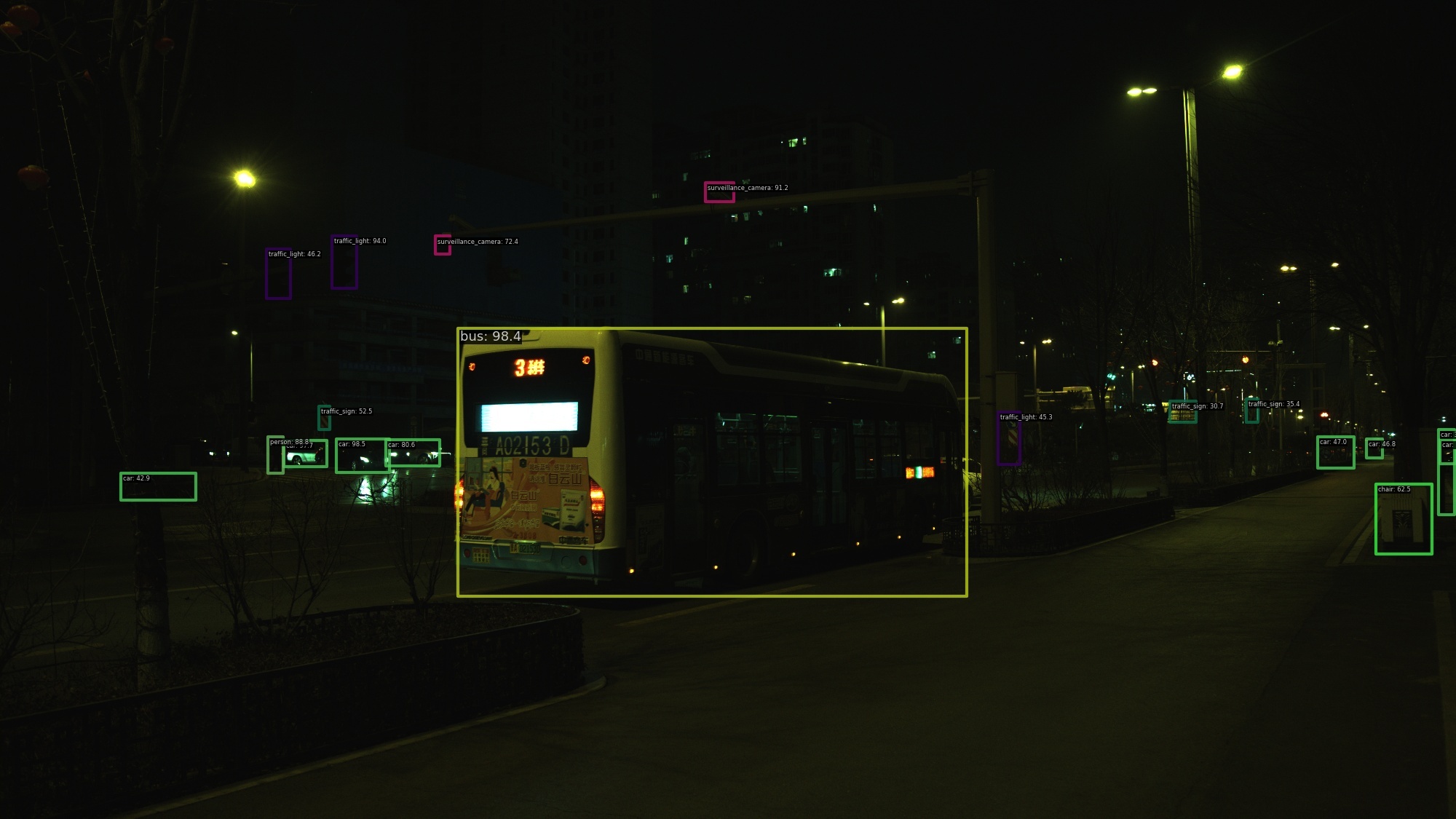}
		\caption{RAW-Adapter}
	\end{subfigure}
	\begin{subfigure}[b]{0.24\linewidth}
		\includegraphics[width=\linewidth]{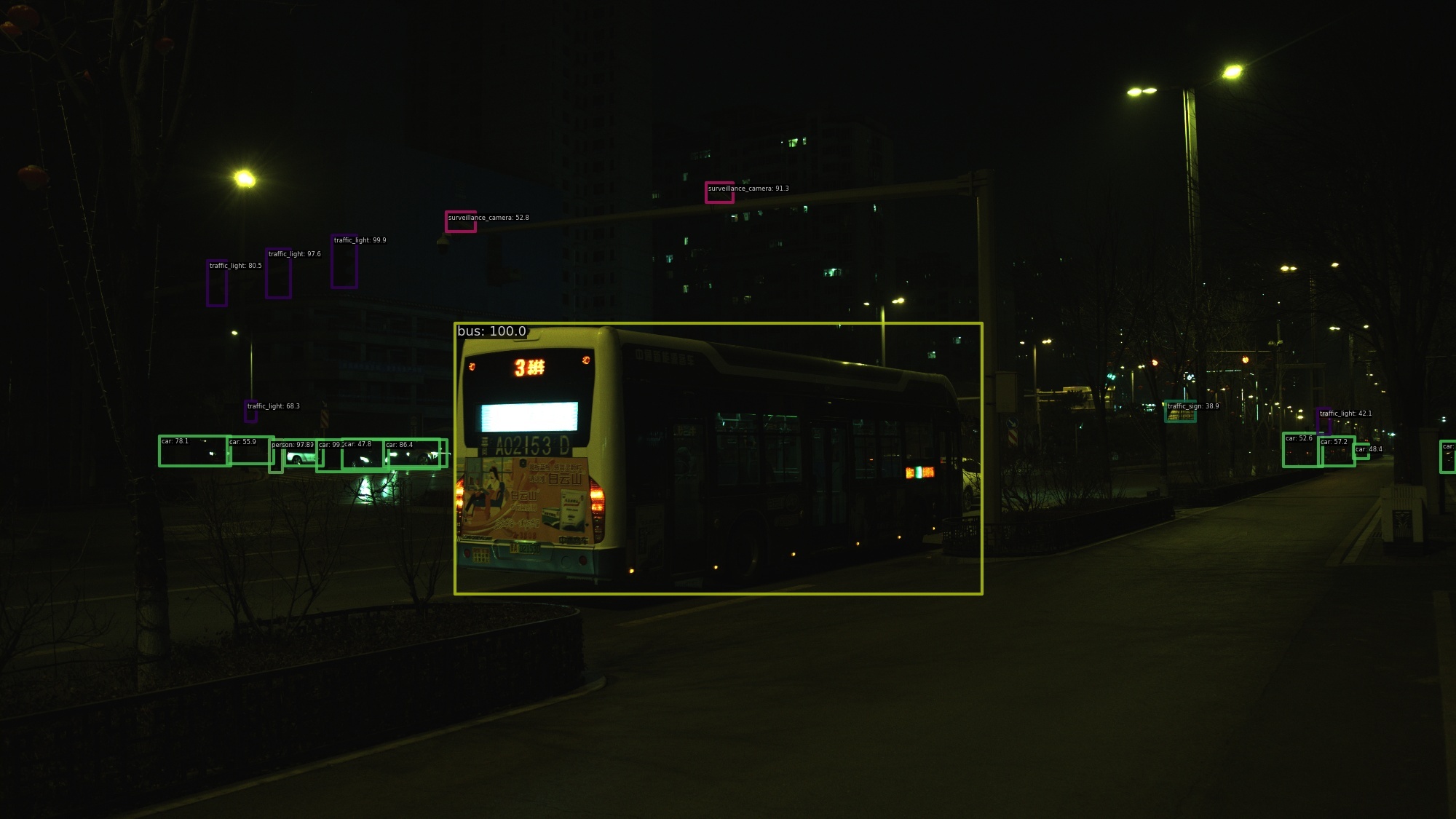}
		\caption{ILLRAW (Ours)}
	\end{subfigure}
	\vspace{-0.5em}
	\caption{Visual examples of detection results with all the comparison methods. Please zoom in for confidence scores and class predictions.}
	\label{fig:visual}
	\vspace{-1em}
\end{figure*}
\begin{figure}
    \vspace{-1em}
	\centering
	\begin{subfigure}{0.43\linewidth}
		\includegraphics[width=\linewidth]{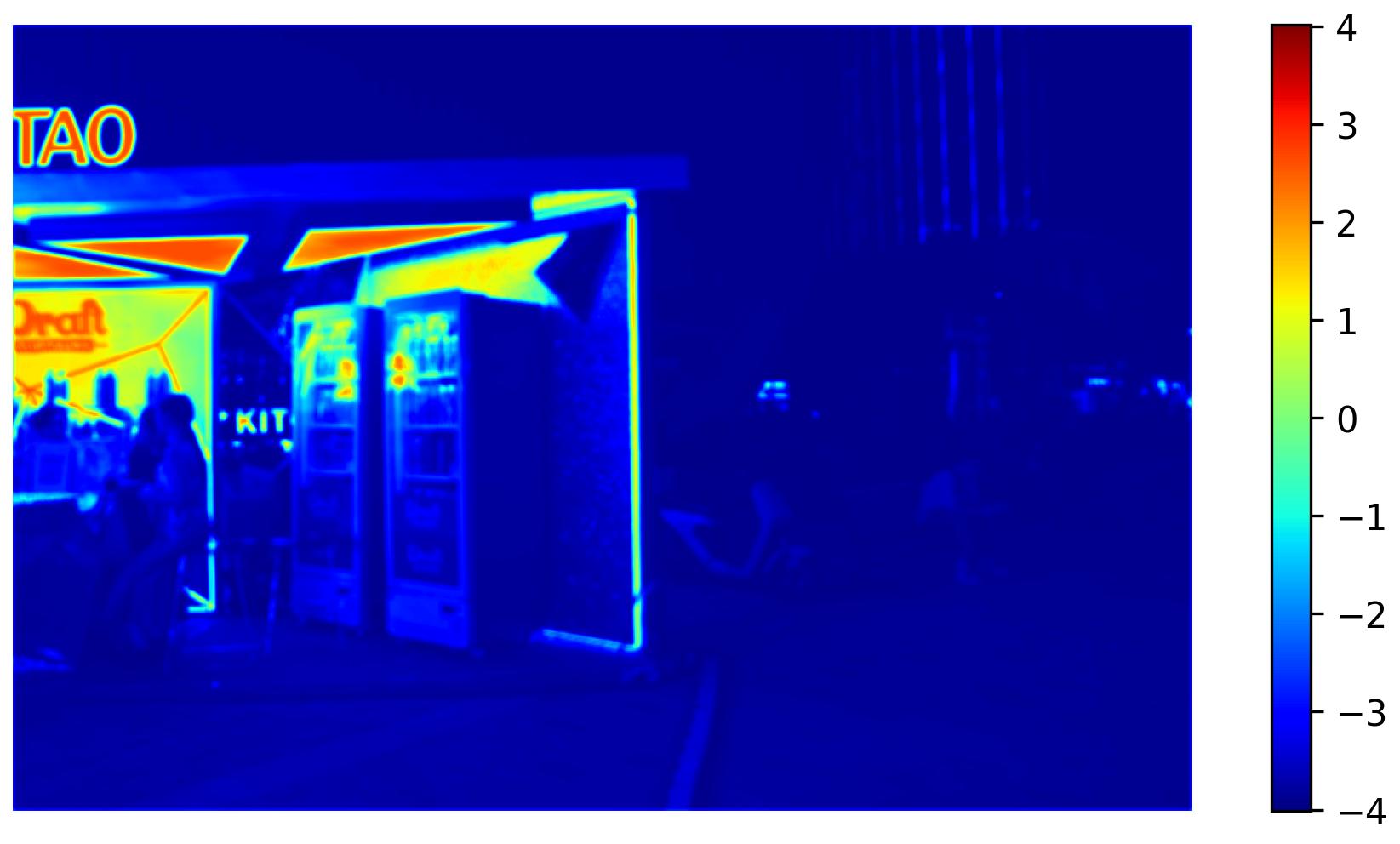}
		\caption{$\mathbf{M}$}
	\end{subfigure}
	\begin{subfigure}{0.43\linewidth}
		\includegraphics[width=\linewidth]{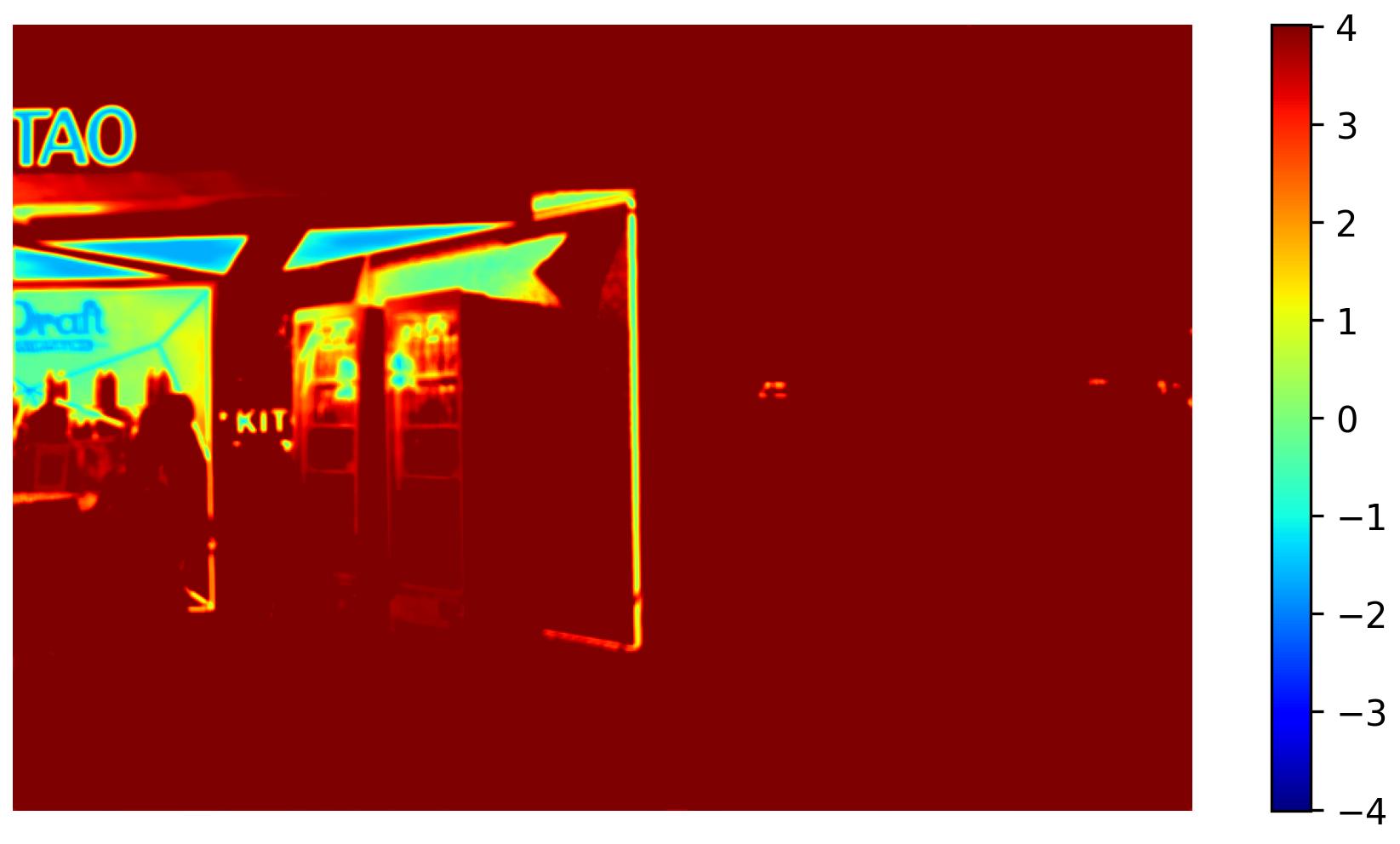}
		\caption{$1-\mathbf{M}$}
	\end{subfigure}
	\vspace{-0.5em}
	\caption{Visualization of the illumination mask $\mathbf{M}$ and its complement $1-\mathbf{M}$. They effectively distinguish different brightness regions in the image, guiding the model to apply distinct operations for areas with varying illumination.}
	\label{fig:ill_mask}
	\vspace{-1.5em}
\end{figure}

\noindent \textbf{Visual Comparison.} 
We further visualize detection results in Figure \ref{fig:visual}. The camera-generated sRGB image shows higher visual quality, with better color reproduction and sharper edges. However, it fails to detect most small objects, such as traffic lights and surveillance cameras. In contrast, most RAW-based methods detect the majority of these small objects. The two open-source ISP methods, Karaimer \textit{et al.} \cite{karaimer2016software} and InvISP \cite{xing2021invertible}, improve image brightness and enhance recognition in dark regions. However, their processing changes spatial information and leads to several false positives, such as trees misclassified as traffic cones and vehicles incorrectly detected near image borders. The RAW-based detectors RAOD \cite{xu2023toward} and RAW-Adapter \cite{cui2024raw} show promising results but still fail to handle dark regions effectively; several objects in the darker left area remain undetected. Our method recalls some objects missing from the dataset annotations, such as a traffic light in the left dark region and distant vehicles. These objects are inherently difficult to detect, and this result highlights the robustness of our method rather than annotation errors. Figure \ref{fig:ill_mask} shows the estimated illumination mask $\mathbf{M}$ and its complement $1-\mathbf{M}$. These masks clearly separate regions with different brightness levels and guide the model to apply illumination-aware processing.

\subsection{Ablation Studies}
\label{sec:ablation}

In this section, we perform ablation studies to validate the effectiveness of the proposed model components. Experiments are conducted on the AODRaw dataset using Cascade-RCNN as the detector.

    \noindent \textbf{Effects of CSE.}
We first assess the contribution of CSE. 
The RAW baseline of 18.1 mAP in Table~\ref{tab:aodraw} is reported by~\cite{li2024towards} using gamma-corrected RAW input. In contrast, the 16.8 mAP baseline in Table~\ref{tab:ablation} is retrained without gamma correction; ILLRAW and all ablation variants follow this same setting. As presented in Table \ref{tab:ablation}, introducing either the green (G) or red/blue (RB) guidance branches to the baseline significantly improves performance. Furthermore, combining both guidance branches achieves the best results, even when using a simple summation for fusion, yielding an improvement of 5.2 mAP.
\begin{table}
	\setlength{\tabcolsep}{5.5pt}
	\centering
    \vspace{-1em}
    \caption{Ablation study on AODRaw without gamma correction.}
	\label{tab:ablation}
		\begin{tabular}{ccccccccc}
			\toprule
			\multicolumn{2}{c}{CSE} && \multicolumn{4}{c}{Fusion module} & &\multirow{2}*{mAP} \\
			\cmidrule{1-2}
			\cmidrule{4-7}
			G & RB && Mask & Sum & Cat & IAF &&  \\
			\midrule
			- & - && - & - & - & - && 16.8 \\
			\checkmark & - && - & - & - & - && 20.4 \\
			- &  \checkmark && - & - & - & - && 21.1 \\
			\midrule
			\checkmark & \checkmark && - & \checkmark & - & - && 22.0 \\
			\checkmark & \checkmark && \checkmark & \checkmark & - & - &&  22.1 \\
			\checkmark & \checkmark && - & - & \checkmark & - && 22.4 \\
			\checkmark & \checkmark && - & - & - & \checkmark && 22.8 \\
			\checkmark & \checkmark && \checkmark & - & - & \checkmark && 23.3  \\
			\bottomrule
		\end{tabular}
	\vspace{-1em}
\end{table}

\noindent \textbf{Effects of IAF.}
We investigate the benefits of the proposed IAF by comparing three fusion strategies. These strategies are direct feature summation, weighted summation guided by the illumination mask, and concatenation followed by convolutional aggregation. Given that the illumination mask plays a key role in guiding feature fusion within the IAF module, we conduct an ablation study to evaluate its impact alongside the fusion methods. For the ``IAF on, Mask off'' setting, the query is generated from the embedded features $\mathbf{Y}_{RB}'$ and $\mathbf{Y}_{G}'$ instead of $\mathbf{M}$, and neither value branch is modulated by $\mathbf{M}$ or $1-\mathbf{M}$; all key/value projections and attention operations remain unchanged. As shown in Table \ref{tab:ablation}, all three fusion strategies achieve comparable accuracy. Weighted summation improves mAP by 0.1, enhancing the model’s sensitivity to bright regions. Concatenation-based fusion offers greater flexibility, resulting in 22.4 mAP. Our proposed IAF module surpasses concatenation by 0.9 mAP, reaching 23.3 mAP. This gain is attributed to the IAF’s dynamic utilization of the illumination mask for attention-guided fusion. Furthermore, removing the illumination mask from our method causes a 0.5 mAP drop, underscoring the critical role of the illumination mask in fusion.


\begin{wraptable}{r}{5.5cm}
	\centering
	\vspace{-2em} 
	\caption{Impact of backbone pretraining on our method.}
	\label{tab:pretrain}
	\setlength{\tabcolsep}{4pt} 
	\resizebox{\linewidth}{!}{%
	\begin{tabular}{lccc} 
		\toprule
		Method & mAP & AP$_{50}$ & AP$_{75}$ \\
		\midrule
		Random & 21.7 & 34.7 & 22.8 \\
		ResNet-18 & 22.4 & 36.1 & 24.0 \\
		sRGB & 23.0 & \textbf{36.2} & 24.7 \\
		Synthesized RAW & \textbf{23.3} & 36.0 & \textbf{24.8} \\
		\bottomrule
	\end{tabular}
	}
	\vspace{-1em} 
\end{wraptable}
\vspace{0.5em}
\noindent \textbf{Impact of pretrained weights.} We analyze the impact of backbone pretraining on our method using a ResNet-18 backbone. Pretrained weights usually improve detector performance, but our modification of the first layer reduces the direct applicability of standard ImageNet weights. We compare four initialization strategies. These strategies are random initialization, ResNet-18 pretrained weights, ILLRAW pretrained on ImageNet in the sRGB domain, and ILLRAW pretrained on ImageNet with synthesized RAW images
\cite{brooks2019unprocessing}. The results in Table \ref{tab:pretrain} show that our method achieves competitive performance without pretrained weights. Using ResNet pretrained weights provides a small improvement to 22.4 mAP. Pretraining ILLRAW on sRGB images adds another 0.6 mAP, which compensates for the missing weights in the modified layer. Pretraining on synthesized RAW images gives the best performance at 23.3 mAP, likely because the pretraining domain matches RAW data and improves illumination modeling and noise robustness.


\section{Conclusion}
In this paper, we propose a novel illumination-aware RAW object detection method, addressing the unique challenges posed by the inherent properties of RAW images. Unlike prior approaches that rely on human visual priors to design learnable ISPs, our method, ILLRAW, leverages the intrinsic variations in light intensity perception across both channel and spatial dimensions of RAW data. We introduce an innovative illumination mask estimation module alongside a channel-sensitive embedding strategy, effectively capturing and processing the semantic information embedded within RAW images. Furthermore, an illumination attention fusion mechanism adaptively enhances feature integration across regions with varying brightness levels.  
Collectively, these contributions significantly advance the accuracy and efficiency of RAW-based object detection. 

\subsubsection{Acknowledgments.} This work was funded by the China Postdoctoral Science Foundation 2025M784248.

\bibliographystyle{splncs04}
\bibliography{main}
\end{document}